\documentclass[twocolumn]{article}

\usepackage[utf8]{inputenc}
\usepackage{amsmath,amsthm,amssymb,amsfonts}
\usepackage{fullpage}
\usepackage{microtype}
\usepackage[round]{natbib}
\usepackage{enumerate}
\usepackage{graphicx,subfigure}
\usepackage{algorithm,algorithmic}
\usepackage{hyperref}
\usepackage{booktabs}

\def\map{\textsc{map}}
\def\bo{\boldsymbol\omega}
\def\bt{{\boldsymbol\theta}}
\def\K{\mathbf{K}}

\def\E{\mathbb{E}}
\def\tU{\widetilde{U}}
\def\N{\mathcal{N}}
\def\X{\mathcal{X}}
\def\dX{\dot{\mathcal{X}}}

\def\L{\mathcal{L}}
\def\bL{\mathbf{L}}
\def\bl{\boldsymbol\ell}

\def\y{\mathbf{y}}
\def\u{\mathbf{u}}
\def\z{\mathbf{z}}

\def\f{\mathbf{f}}

\def\x{\mathbf{x}}

\def\dx{\dot{\mathbf{x}}}
\def\0{\mathbf{0}}
\def\cov{\mathrm{cov}}

\def\diag{\mathrm{diag}}

\def\R{\mathbb{R}}
\def\GP{\mathcal{GP}}

\def\vdp{\textsc{vdp}}
\def\fhn{\textsc{fhn}}
\def\lv{\textsc{lv}}

\DeclareMathOperator*{\argmax}{arg\,max}

\newcommand{\gN}{\mathcal{N}}

\title{Learning unknown ODE models with Gaussian processes}

\author{Markus Heinonen$^{1,2,*}$ \and Cagatay Yildiz$^{1,}$\thanks{Equal contribution} \and Henrik Mannerström$^1$ \and Jukka Intosalmi$^1$ \and Harri Lähdesmäki$^1$}

\date{$^1$ Department of Computer Science, Aalto university\\$^2$ Helsinki Institute for Information Technology HIIT}

\begin{document} 

\maketitle

\vspace{20mm}

\begin{abstract}
In conventional ODE modelling coefficients of an equation driving the system state forward in time are estimated. However, for many complex systems it is practically impossible to determine the equations or interactions governing the underlying dynamics. In these settings, parametric ODE model cannot be formulated. Here, we overcome this issue by introducing a novel paradigm of nonparametric ODE modeling that can learn the underlying dynamics of arbitrary continuous-time systems without prior knowledge. We propose to learn non-linear, unknown differential functions from state observations using Gaussian process vector fields within the exact ODE formalism. We demonstrate the model's capabilities to infer dynamics from sparse data and to simulate the system forward into future.
\end{abstract}

\section{Introduction}

Dynamical systems modeling is a cornerstone of experimental sciences. In biology, as well as in physics and chemistry, modelers attempt to capture the dynamical behavior of a given system or a phenomenon in order to improve its understanding and make predictions about its future state. Systems of coupled ordinary differential equations (ODEs) are undoubtedly the most widely used models in science. Even simple ODE functions can describe complex dynamical behaviours \citep{Hirsch04}. Typically, the dynamics are firmly grounded in physics with only a few parameters to be estimated from data. However, equally ubiquitous are the cases where the governing dynamics are partially or completely unknown. 

We consider the dynamics of a system governed by multivariate ordinary differential functions:
\begin{align} \label{eq:f}
\dx(t) = \frac{d\x(t)}{dt} = \f(\x(t))
\end{align}
where $\x(t) \in \X = \mathbb{R}^D$ is the state vector of a $D$-dimensional dynamical system at time $t$,
and the $\dx(t) \in \dX = \R^D$ is the first order time derivative of $\x(t)$ that drives the state $\x(t)$ forward, and where $\f : \R^D \rightarrow \R^D$ is the vector-valued derivative function. The ODE solution is determined by
\begin{align} \label{eq:ode}
\x(t) &= \x_0 + \int_0^t \f(\x(\tau)) d\tau,
\end{align}
where we integrate the system state from an initial state $\x(0) = \x_0$ for time $t$ forward. We assume that $\f(\cdot)$ is \emph{completely unknown} and we only observe one or several multivariate time series $Y = (\y_1, \ldots, \y_N)^T \in \R^{N \times D}$ obtained from an additive noisy observation model at  observation time points $T = (t_1, \ldots, t_N) \in \R^N$,
\begin{equation}
\y(t) = \x(t) + \varepsilon_t,
\end{equation}
where $\varepsilon_t \sim \N(\0, \Omega)$ follows a stationary zero-mean multivariate Gaussian distribution with diagonal noise variances $\Omega = \diag( \omega_1^2, \ldots, \omega_D^2)$. The observation time points do not need to be equally spaced. Our task is to learn the differential function $\f(\cdot)$ given observations $Y$, with no prior knowledge of the ODE system.

There is a vast literature on conventional ODEs \citep{butcher2016} where a parametric form for function $\f(\x ; \bt, t)$ is assumed to be known, and its parameters $\bt$ are subsequently optimised with least squares or Bayesian approach, where the expensive forward solution $\x_\bt(t_i) = \int_0^{t_i} \f(\x(\tau) ; \bt, t) d\tau$ is required to evaluate the system responses $\x_\bt(t_i)$ from parameters $\bt$ against observations $\y(t_i)$. To overcome the computationally intensive forward solution, a family of methods denoted as gradient matching \citep{Varah1982,Ellner2002,ramsay2007} have proposed to replace the forward solution by matching $\f(\y_i) \approx \dot{\y}_i$ to empirical gradients $\dot{\y}_i$ of the data instead, which do not require the costly integration step. Recently several authors have proposed embedding a parametric differential function within a Bayesian or Gaussian process (GP) framework \citep{graepel2003,Calderhead2008,dondelinger2013,wang2014,macdonald2017} (see \citet{macdonald2015} for a review). GPs have been successfully applied to model linear differential equations as they are analytically tractable \citep{gao2008,raissi2017}. 

However, conventional ODE modeling can only proceed if a parametric form of the driving function $\f(\cdot)$ is known. Recently, initial work to handle unknown or non-parametric ODE models have been proposed, although with various limiting approximations. Early works include spline-based smoothing and additive functions $\sum_j^D f_j(x_j)$ to infer gene regulatory networks \citep{dehoon2003,henderson2014}. \citet{aijo2009} proposed estimating the unknown nonlinear function with GPs using either finite time differences, or analytically solving the derivative function as a function of only time, $\dot{\x}(t) = \f(t)$ \citep{aijo2013}. In a seminal technical report of \citet{heinonen2014} a full vector-valued kernel model $\f(\x)$ was proposed, however using a gradient matching approximation. To our knowledge, there exists no model that can learn non-linear ODE functions $\dx(t) = \f(\x(t))$ over the state $\x$ against the true forward solutions $\x(t_i)$. 

In this work we propose \textsc{npODE}: the first ODE model for learning arbitrary, and \emph{a priori} completely unknown non-parametric, non-linear differential functions $\f : \X \rightarrow \dX$ from data in a Bayesian way. We do not use gradient matching or other approximative models, but instead propose to directly optimise the exact ODE system with the fully forward simulated responses against data. We parameterise our model as an augmented Gaussian process vector field with inducing points, while we propose sensitivity equations to efficiently compute the gradients of the system. Our model can forecast continuous-time systems arbitrary amounts to future, and we demonstrate the state-of-the-art performance in human motion datasets. The MATLAB implementation is publicly available at \url{github.com/cagatayyildiz/npode}.

\section{Nonparametric ODE model}

The differential function $\f(\x)$ to be learned defines a \emph{vector field}\footnote{We use vector field and differential function interchangeably.} $\f$, that is, an assignment of a gradient vector $\f(\x) \in \R^D$ to every state $\x \in \R^D$. We model the vector field as a vector-valued Gaussian process (GP) \citep{rasmussen2006}
\begin{align}
\f(\x) &\sim \GP( \0, K(\x,\x')),
\end{align}
which defines \emph{a priori} distribution over function values $\f(\x)$ whose mean and covariances are
\begin{align}
\E[ \f(\x)] &= \0 \\
\cov[ \f(\x), \f(\x')] &= K(\x,\x'),
\end{align}
and where the kernel $K(\x,\x') \in \R^{D \times D}$ is matrix-valued. A GP prior defines that for any collection of states $X = (\x_1, \ldots, \x_N)^T \in \R^{N \times D}$, the function values $F = ( \f(\x_1), \ldots, \f(\x_N))^T \in \R^{N \times D}$ follow a matrix-valued normal distribution,  
\begin{align}
p(F) = \N(vec(F) | \0, \K(X,X)),
\end{align}
where $\K(X,X) = (K(\x_i, \x_j))_{i,j=1}^N \in \R^{ND \times ND}$ is a block matrix of matrix-valued kernels $K(\x_i,\x_j)$. The key property of Gaussian processes is that they encode functions where similar states $\x,\x'$ induce similar differentials $\f(\x),\f(\x')$, and where the state similarity is defined by the kernel $K(\x,\x')$. 

\begin{figure}[t]
    \centering
    \includegraphics[width=0.99\columnwidth]{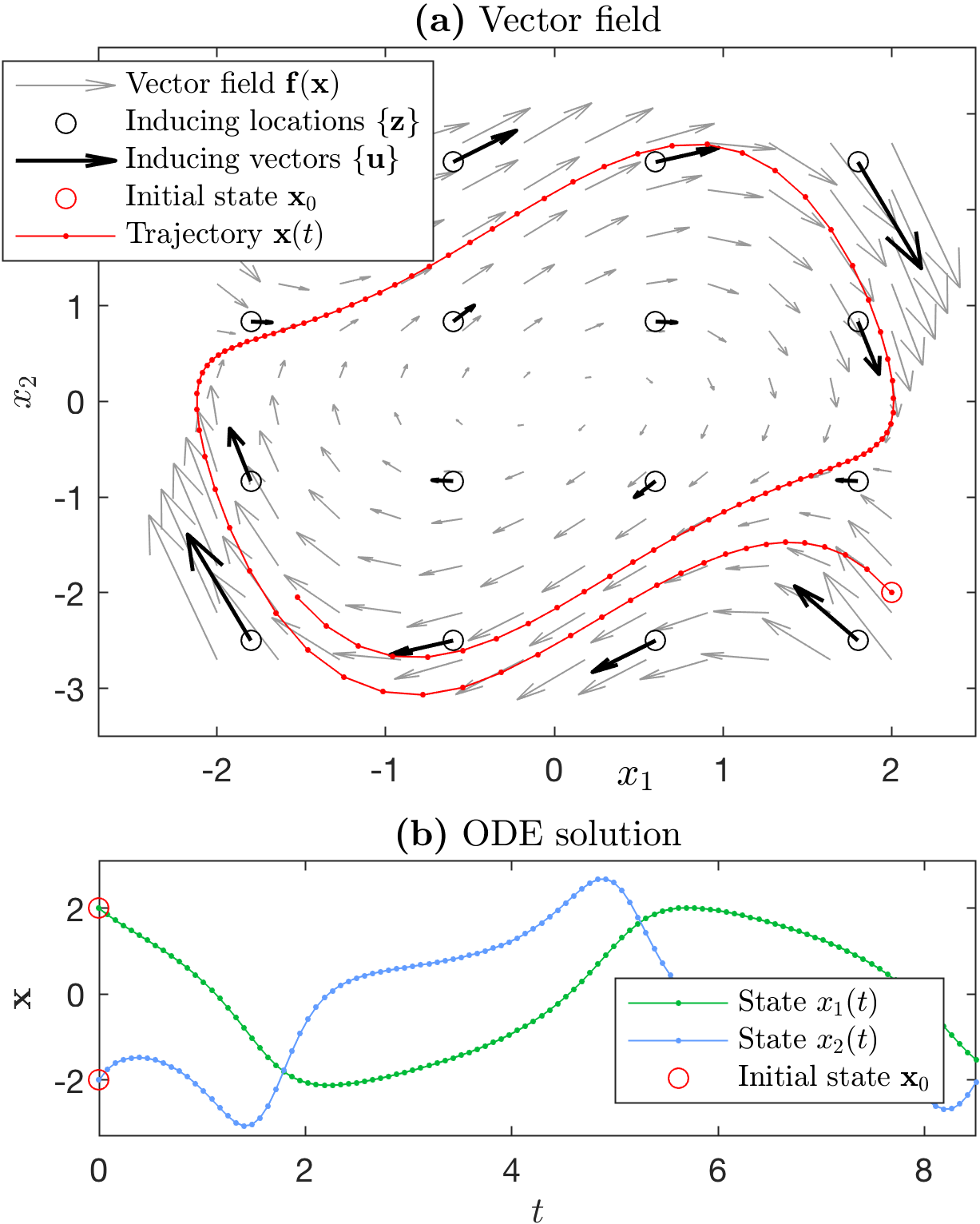}
    \caption{\textbf{(a)} Illustration of an ODE system vector field induced by the Gaussian process. The vector field $\f(\x)$ (gray arrows) at arbitrary states $\x$ is interpolated from the inducing points $\u,\z$ (black arrows), with the trajectory $\x(t)$ (red points) following the differential system $\f(\x)$ exactly. \textbf{(b)} The trajectory $\x(t)$ plotted over time $t$.}
    \label{fig:ode}
\end{figure}

In standard GP regression we would obtain posterior of the vector field by conditioning the GP prior with the data \citep{rasmussen2006}. In ODE models the conditional $\f(\x) | Y$ of a vector field is intractable due to the integral mapping \eqref{eq:ode} between observed states $\y(t_i)$ and differentials $\f(\x)$. Instead, we resort to augmenting the Gaussian process with a set of $M$  \emph{inducing points} $\z \in \X$ and $\u \in \dX$, such that $\f(\z) = \u$ \citep{quinonero2005}. We choose to interpolate the differential function between the inducing points as (See Figure \ref{fig:ode})
\begin{align}
\f(\x) &\triangleq \K_\bt(\x, Z) \K_\bt(Z,Z)^{-1} vec(U), \label{eq:vf}
\end{align}
which supports the function $\f(\x)$ with \emph{inducing locations} $Z = (\z_1, \ldots, \z_M)$, \emph{inducing vectors} $U = (\u_1, \ldots, \u_M)$, and $\bt$ are the kernel parameters. The function above corresponds to a vector-valued kernel function \citep{alvarez2012}, or to a multi-task Gaussian process conditional mean without the variance term \citep{rasmussen2006}. This definition is then compatible with the deterministic nature of the ODE formalism. Due to universality of several kernels and kernel functions \citep{shawe2004kernel}, we can represent arbitrary vector fields with appropriate inducing point and kernel choices.

\subsection{Operator-valued kernels}
\label{sec:rbf}

The vector-valued kernel function \eqref{eq:vf} uses \emph{operator-valued} kernels, which result in matrix-valued kernels $K_\bt(\z,\z') \in \R^{D \times D}$ for real valued states $\x,\z$, while the kernel matrix over data points becomes $\K_\bt = ( K(\z_i, \z_j))_{i,j=1}^M \in \R^{MD \times MD}$ (See \citet{alvarez2012} for a review). Most straightforward operator-valued kernel is the identity decomposable kernel $K_{dec}(\z,\z') = k(\z,\z') \cdot I_D$, where the scalar Gaussian kernel
\begin{align}
K_\bt(\z,\z') = \sigma_f^2 \exp \left(- \frac{1}{2} \sum_{j=1}^D \frac{ (z_j - z_j')^2}{\ell_j^2} \right)
\end{align}
with differential variance $\sigma_f^2$ and dimension-specific lengthscales $\bl = (\ell_1, \ldots, \ell_D)$ is expanded into a diagonal matrix of size $D \times D$. We collect the kernel parameters as $\bt = (\sigma_f, \bl)$.

We note that more complex kernels can also be considered given prior information of the underlying system characteristics. The divergence-free matrix-valued kernel induces vector fields that have zero divergence \citep{wahlstrom2013,solin2015}. Intuitively, these vector fields do not have sinks or sources, and every state always finally returns to itself after sufficient amount of time. Similarly, curl-free kernels induce curl-free vector fields that can contain sources or sinks, that is, trajectories can accelerate or decelerate. For theoretical treatment of vector field kernels, see \citep{narcowich1994generalized,bhatia2013,fuselier2017}. Non-stationary vector fields can be modeled with input-dependent lengthscales \citep{heinonen2016}, while spectral kernels can represent stationary \citep{wilson2013} or non-stationary \citep{remes2017} recurring patterns in the differential function.

\subsection{Joint model}

We assume a Gaussian likelihood over the observations $\y_i$ and the corresponding simulated responses $\x(t_i)$ of Equation \eqref{eq:ode}, 
\begin{align}
p(Y | \x_0, U, Z, \bo) &= \prod_{i=1}^N \N(\y_i | \x(t_i), \Omega),
\end{align}
where $\x(t_i)$ are forward simulated responses using the integral equation \eqref{eq:ode} and differential equation \eqref{eq:vf}, and $\Omega = \diag( \omega_1^2 \ldots, \omega_D^2)$ collects the dimension-specific noise variances. 

The inducing vectors have a Gaussian process prior
\begin{align}
p(U | Z, \bt) &= \N( vec(U) | \0, \K_\bt(Z,Z)).
\end{align}
The model posterior is then
\begin{align} 
p(U,\x_0,\bt,\bo | Y) &\propto p(Y | \x_0, U,\bo) p(U|\bt) = \mathcal{L}, \label{eq:post}
\end{align}
where we have for brevity omitted the dependency on the locations of the inducing points $Z$ and also the parameter hyperpriors $p(\bt)$ and $p(\bo)$ since we assume them to be uniform, unless there is specific domain knowledge of the priors. 

The model parameters are the initial state $\x_0$\footnote{In case of multiple time-series, we will use one initial state for each time-series.}, the inducing vectors $U$, the noise standard deviations $\bo = (\omega_1, \ldots, \omega_D)$, and the kernel hyperparameters $\bt = (\sigma_f, \ell_1, \ldots, \ell_D)$. 


\subsection{Noncentral parameterisation} 

We apply a latent parameterisation using Cholesky decomposition $\bL_\bt \bL_\bt^T = \K_\bt(Z,Z)$, which maps the inducing vectors to whitened domain \citep{kuss2005}
\begin{align}
U &= \bL_\bt \tU, \qquad \tU = \bL_\bt^{-1} U.
\end{align}
The latent variables $\tU$ are projected on the kernel manifold $\bL_\bt$ to obtain the inducing vectors $U$. This non-centered parameterisation (NCP) transforms the hierarchical posterior $\mathcal{L}$ of Equation~\eqref{eq:post} into a reparameterised form \begin{align} 
p(\x_0,\tU,\bt,\bo | Y) &\propto p(Y | \x_0, \tU, \bo,\bt) p(\tU), \label{eq:ncp}
\end{align}
where all variables to be optimised are decoupled, with the latent inducing vectors having a standard normal prior $\tU \sim \N(\0,I)$. Optimizing $\tU$ and $\bt$ is now more efficient since they have independent contributions to the vector field via $U = \bL_\bt \tU$. 
The gradients of the whitened posterior can be retrieved analytically as \citep{heinonen2016}
\begin{align}
\nabla_{\tU} \log \mathcal{L} = \bL_\bt^T \nabla_U \log \L. \label{eq:white}
\end{align}

Finally, we find a MAP estimate for the initial state $\x_0$, latent vector field $\tU$, kernel parameters $\bt$ and noises $\bo$ by gradient ascent,
\begin{align}
\x_{0,\map}, \tU_\map, \bt_\map, \bo_\map &= \argmax_{\x_0,\tU,\bt,\bo} \log \L,
\end{align}
while keeping the inducing locations $Z$ fixed on a sufficiently dense grid (See Figure \ref{fig:ode}). The partial derivatives of the posterior with respect to noise parameters $\bo$ can be found analytically, while the derivatives with respect to $\sigma_f$ are approximated with finite differences. We select the optimal lengthscales $\bl$ by cross-validation.

\section{Sensitivity equations}

The key term to carry out the MAP gradient ascent optimization is the likelihood
$$\log p(Y | \x_0, \tU, \bo)$$
that requires forward integration and computing the partial derivatives with respect to the whitened inducing vectors $\tU$. Given Equation \eqref{eq:white} we only need to compute the gradients with respect to the inducing vectors $\u = vec(U) \in \R^{MD}$,
\begin{align}
&\frac{d \log p(Y | \x_0, \u, \bo)}{d \u} \notag \\
  &= \sum_{s=1}^{N} \frac{d \log \N(\y_s | \x(t_s,\u), \Omega)}{d\x}\frac{d\x(t_s,\u)}{d \u}.
\end{align}
This requires computing the derivatives of the simulated system response $\x(t,\u)$ against the vector field parameters $\u$,
\begin{align}
\frac{d \x(t, \u)}{d \u} \equiv S(t) \in \R^{D \times MD},
\end{align}
which we denote by $S_{ij}(t) = \frac{\partial \x(t, \u)_i}{\partial u_j}$, and expand the notation to make the dependency of $\x$ on $\u$ explicit. Approximating these with finite differences is possible in principle, but is highly inefficient and has been reported to cause unstability \citep{raue2013}. We instead turn to sensitivity equations for $\u$ and $\x_0$ that provide computationally efficient, analytical gradients $S(t)$ \citep{kokotovic1967,froechlich2017}.

The solution for $\frac{d\x(t,\u)}{d \u}$ can be derived by differentiating the full nonparametric ODE system with respect to $\u$ by
\begin{align}
\frac{d}{d\u} \frac{d\x(t,\u)}{dt}  &= \frac{d}{d\u} \f(\x(t,\u)).
\end{align}
The sensitivity equation for the given system can be obtained by changing the order of differentiation on the left hand side and carrying out the differentiation on the right hand side. The resulting sensitivity equation can then be expressed in the form
\begin{align}
\overbrace{\frac{d}{dt} \frac{d\x(t,\u)}{d\u}}^{\dot{S}(t)} = \overbrace{\frac{\partial \f(\x(t,\u))}{\partial \x}}^{J(t)}  \overbrace{\frac{d\x(t,\u)}{d\u}}^{S(t)} + \overbrace{\frac{\partial \f(\x(t,\u))}{\partial \u}}^{R(t)}, \label{sens_eq}
\end{align}
where $J(t) \in \R^{D \times D}$, $R(t), \dot{S}(t) \in \R^{D \times MD}$ (See Appendix for detailed specification).
For our nonparametric ODE system the sensitivity equation is fully determined by
\begin{align}
J(t) &= \frac{\partial \K(\x,Z)}{\partial \x} \K(Z,Z)^{-1} \u  \\
R(t) &= \K(\x,Z) \K(Z,Z)^{-1}.
\end{align}

The sensitivity equation provides us with an additional ODE system which describes the time evolution of the derivatives with respect to the inducing vectors $S(t)$. The sensitivities are coupled with the actual ODE system and, thus, both systems $\x(t)$ and $S(t)$ are concatenated as the new augmented state that is solved jointly by Equation~\eqref{eq:ode} driven by the differentials $\dot{\x}(t)$ and $\dot{S}(t)$ \citep{leis1988}. The initial sensitivities are computed as $S(0) = \frac{d \x_0}{d \u}$. In our implementation, we merge $\x_0$ with $\u$ for sensitivity analysis to obtain the partial derivatives with respect to the initial state which is estimated along with the other parameters. We use the \textsc{cvodes} solver from the \textsc{Sundials} package \citep{hindmarsh2005} to solve the nonparametric ODE models and the corresponding gradients numerically. The sensitivity equation based approach is superior to the finite differences approximation because we have exact formulation for the gradients of state over inducing points, which can be solved up to the numerical accuracy of the ODE solver.

\section{Simple simulated dynamics}
\label{sec:simple}

\begin{figure*}[t]
     \centering
     \includegraphics[width=2.08\columnwidth]{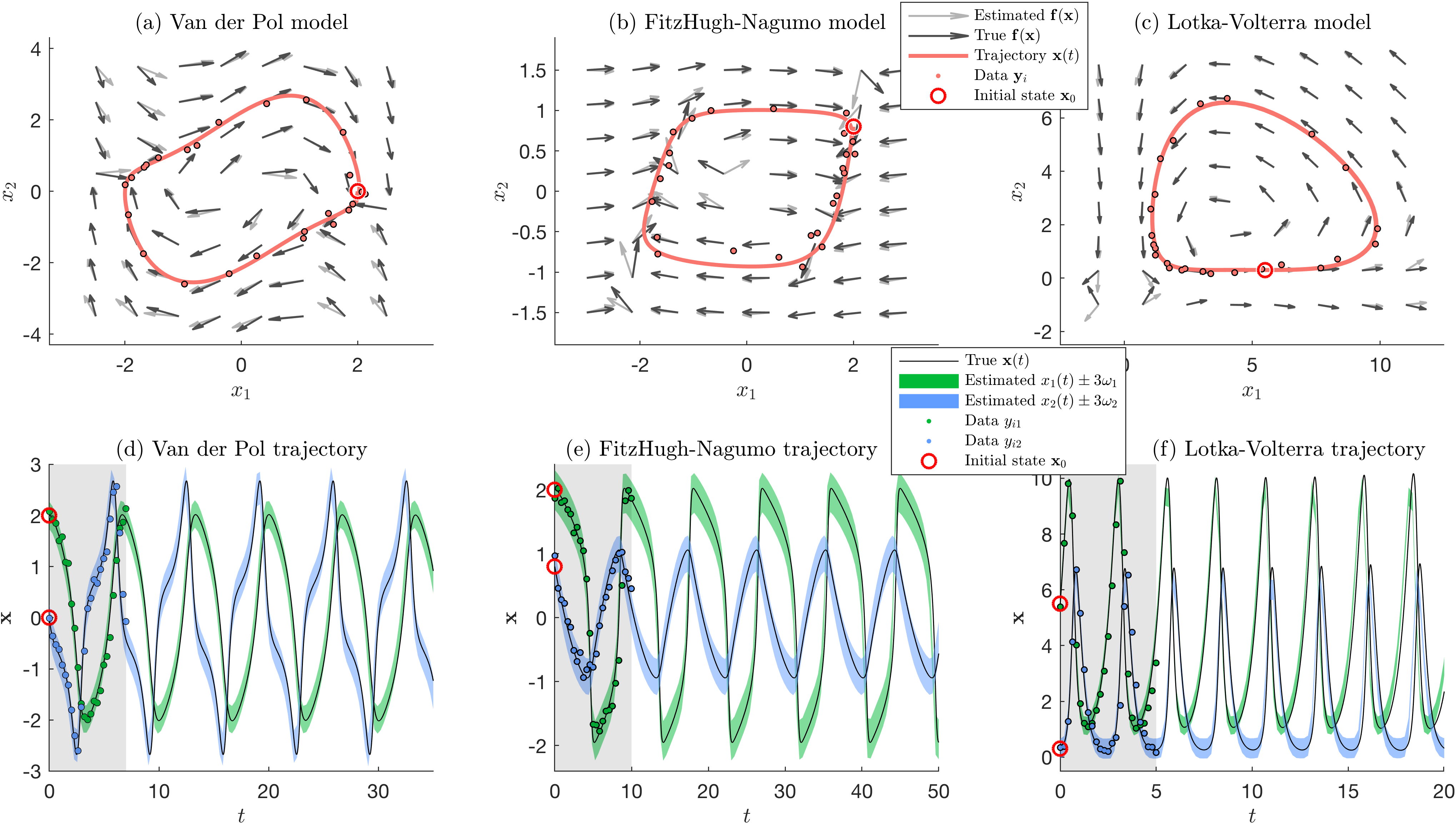}
     \caption{Estimated dynamics from Van der Pol, FitzHugh-Nagumo and Lotka-Volterra systems. The top part \textbf{(a-c)} shows the learned vector field (grey arrows) against the true vector field (black arrows). The bottom part \textbf{(d-f)} shows the training data (grey region points) and forecasted future cycle likelihoods with the learned model (shaded region) against the true trajectory (black line).}
     \label{fig:osc}
\end{figure*}

As first illustration of the proposed nonparametric ODE method we consider three simulated differential systems: the Van der Pol (VDP), FitzHugh-Nagumo (FHN) and Lotka-Volterra (LV) oscillators of form
\begin{align*}
\vdp: & & \dot{x}_1 &= x_2                   & \dot{x}_2 &= (1-x_1^2) x_2 - x_1 \\
\fhn: & & \dot{x}_1 &= 3(x_1 - \frac{x_1^3}{3} + x_2) & \dot{x}_2 &= \frac{0.2 - 3x_1 - 0.2x_2}{3} \\
\lv:  & & \dot{x}_1 &= 1.5 x_1 - x_1 x_2     & \dot{x}_2 &= -3 x_2 + x_1 x_2.
\end{align*}
In the conventional ODE case the coefficients of these equations can be inferred using standard statistical techniques if sufficient amount of time series data is available \citep{girolami2008,raue2013}. Our main goal is to infer \emph{unknown} dynamics, that is, when these equations are unavailable and we instead represent the dynamics with a nonparametric vector field of Equation~\eqref{eq:vf}. We use these simulated models to only illustrate our model behavior against the true dynamics.

We employ $25$ data points from one cycle of noisy observation data from VDP and FHN models, and $25$ data points from $1.7$ cycles from the LV model with noise variance of $\sigma_n^2 = 0.1^2$. We learn the npODE model with these training data using $M = 5^2$ inducing locations on a fixed grid, and forecast between 4 and 8 future cycles starting from true initial state $\x_0$ at time $0$. Figure \ref{fig:osc} (bottom) shows the training datasets (grey regions), initial states, true trajectories (black lines) and the forecasted trajectory likelihoods (colored regions). The model accurately learns the dynamics from less than two cycles of data and can reproduce them reliably into future.

Figure \ref{fig:osc} (top) shows the corresponding true vector field (black arrows) and the estimated vector field (grey arrows). The vector field is a continuous function, which is plotted on a 8x8 grid for visualisation. In general the most difficult part of the system is learning the middle of the loop (as seen in the FHN model), and learning the most outermost regions (bottom left in the LV model). The model learns the underlying differential $\f(\x)$ accurately close to observed points, while making only few errors in the border regions with no data.

\section{Unknown system estimation}

\begin{figure*}[t]
    \centering
    \includegraphics[width=2.08\columnwidth]{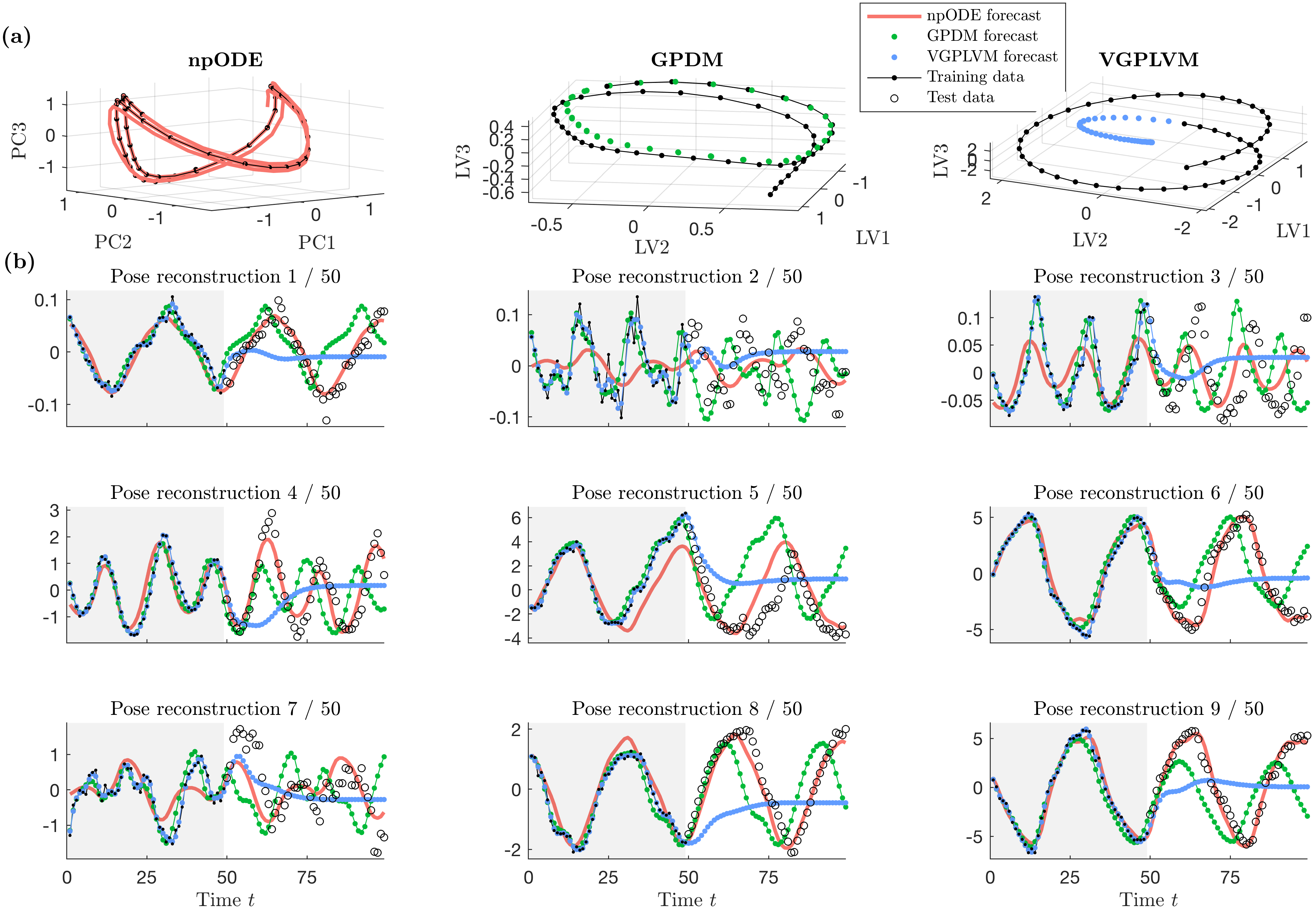}
    \caption{Forecasting 50 future frames after 49 frames of training data of human motion dataset \texttt{35\_12.amc}. \textbf{(a)} The estimated locations of the trajectory in a latent space (black points) and future forecast (colored lines). \textbf{(b)} The original features reconstructed from the latent predictions with grey region showing the training data.}
    \label{fig:forecast}
\end{figure*}

\begin{table}[b]
	\caption{Means and standard deviations of RMSEs of 43 datasets in forecasting and filling experiments.}
	\label{tab:t1}
	\vskip 0.15in
	\begin{center}
    \begin{small}
    \begin{sc}
	\begin{tabular}{lcc}
		\hline
		 Model & Forecasting & Imputation \\ 
	    \hline
		 npODE & $\mathbf{4.52} \pm 2.31$ & $3.94 \pm 3.50$ \\
		 GPDM & $4.94 \pm 3.3$ & $5.31 \pm 3.39$ \\
		 VGPLVM & $8.74 \pm 3.43$ & $\mathbf{3.91} \pm 1.80$ \\
        \hline
    \end{tabular}
    \end{sc}
    \end{small}
    \end{center}
    \vskip -0.1in
\end{table}

Next, we illustrate how the model estimates realistic, unknown dynamics from noisy observations $\y(t_1), \ldots, \y(t_N)$. As in Section~\ref{sec:simple}, we make no assumptions on the structure or form of the underlying system, and capture the underlying dynamics with the nonparameteric system alone. We employ no subjective priors, and assume no inputs, controls or other sources of information. The task is to infer the underlying dynamics $\f(\x)$, and interpolate or extrapolate the state trajectory outside the observed data.

We use a benchmark dataset of human motion capture data from the Carnegie Mellon University motion capture (CMU mocap) database. Our dataset contains $50$-dimensional pose measurements $\y(t_i)$ from humans walking, where each pose dimension records a measurement in different parts of the body during movement \citep{wang2008}. We apply the preprocessing of \citet{wang2008} by downsampling the datasets by a factor of four and centering the data. This resulted in a total of 4303 datapoints spread across $43$ trajectories with on average $100$ frames per trajectory. In order to tackle the problem of dimensionality, we project the original dataset with PCA to a three dimensional latent space where the system is specified, following \citet{damianou2011} and \citet{wang2006}. We place $M=5^3$ inducing vectors on a fixed grid, and optimize our model from 100 initial values, which we select by projecting empirical differences $\y(t_i) - \y(t_{i-1})$ to the inducing vectors. We use an LBFGS optimizer in Matlab. The whole inference takes approximately few minutes per trajectory.

We evaluate the method with two types of experiments: imputing missing values and forecasting future cycles. For the forecasting the first half of the trajectory is for model training, and the second half is to be forecasted. For imputation we remove roughly $20\%$ of the frames from the middle of the trajectory, which are to be filled by the models. We perform model selection for lengthscales $\bl$ with cross-validation split of 80/20. We record the root mean square error (RMSE) over test points in the original feature space in both cases,
where we reconstruct the original dimensions from the latent space trajectories.

Due to the current lack of ODE methods suitable for this nonparametric inference task, we instead compare our method to the state-of-the-art state-space models where such problems have been previously considered  \citep{wang2008}. In a state-space or dynamical model a transition function $\x(t_{k+1}) = \mathbf{g}(\x(t_k))$ moves the system forward in discrete steps. With sufficiently high sampling rate, such models can estimate and forecast finite approximations of smooth dynamics. In Gaussian process dynamical model \citep{wang2006,frigola2014,svensson2016} a GP transition function is inferred in a latent space, which can be inferred with a standard GPLVM \citep{lawrence2004} or with a dependent GPLVM \citep{zhao2016}. In dynamical systems the transition function is replaced by a GP interpolation \citep{damianou2011}. The discrete time state-space models emphasize inference of a low-dimensional manifold as an explanation of the high-dimensional measurement trajectories. 

We compare our method to the dynamical model GPDM of \citet{wang2006} and to the dynamical system VGPLVM of \citet{damianou2011}, where we directly apply the implementations provided by the authors at \url{inverseprobability.com/vargplvm} and \url{dgp.toronto.edu/~jmwang/gpdm}. Both methods optimize their  latent spaces separately, and they are thus not directly comparable.

\subsection{Forecasting}

\begin{figure*}[t]
    \centering
    \includegraphics[width=2.08\columnwidth]{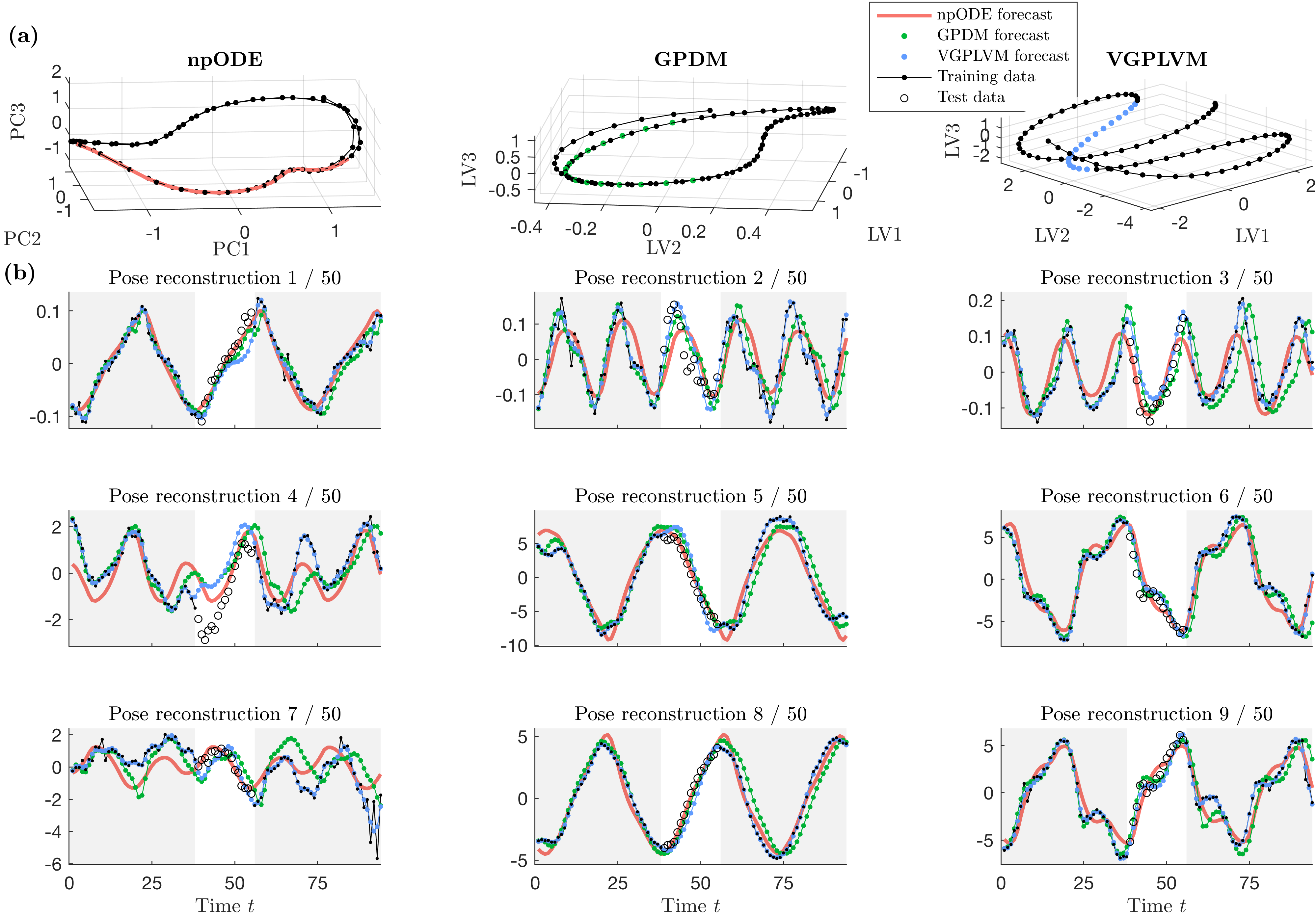}
    \caption{Imputation of 17 missing frames in the middle of a 94-length trajectory of human motion dataset \texttt{07\_07.amc} (subsampled every fourth frame). \textbf{(a)} The estimated locations of the missing points in the latent space are colored. \textbf{(b)} The original features reconstructed from the latent trajectory.}
    \label{fig:impute}
\end{figure*}

In the forecasting task we train all models with the first half of the trajectory, while forecasting the second half starting from the first frame. The models are trained and forecasted within a low-dimensional space, and subsequently projected back into the original space via inverting the PCA or with GPLVM mean predictions. As all methods optimize their latent spaces separately, they are not directly comparable. Thus, the mean errors are computed in the original high-dimensional space. Note that the low-dimensional representation necessarily causes some reconstruction errors.

Figure \ref{fig:forecast} illustrates the models on one of the trajectories \texttt{35\_12.amc}. The top part \textbf{(a)} shows the training data in the PCA space for npODE, and optimized training data representation for GPDM and VGPLVM (black points). The colored lines (npODE) and points (GPDM, VGPLVM) indicate the future forecast. The bottom part \textbf{(b)} shows the first 9 reconstructed original pose dimensions reconstructed from the latent forecasted trajectories. The training data is shown in gray background, while test data is shown with circles. 

The VGPLVM has most trouble forecasting future points, and reverts quickly after training data to a value close to zero, failing to predict future points. The GPDM model produces more realistic trajectories, but fails to predict any of the poses accurately. Finally, npODE can accurately predict five poses, and still retains adequate performance on remaining poses, except for pose 2. 

Furthermore, Table \ref{tab:t1} indicates that npODE is also best performing method on average over the whole dataset in the forecasting.



\subsection{Imputation}

In the imputation task we remove approximately $20\%$ of the training data from the middle of the trajectory. The goal is to learn a model with the remaining data and forecast the missing values. Figure \ref{fig:impute} highlights the performance of the three models on the trajectory \texttt{07\_07.amc}. The top part \textbf{(a)} shows the training data (black points) in the PCA space (npODE) or optimized training locations in the latent space (GPDM, VGPLVM). The middle part imputation is shown with colored points or lines. Interestingly both npODE and GPDM operate on cyclic representations, while VGPLVM is not cyclic.

The bottom panel \textbf{(b)} shows the first 9 reconstructed pose dimensions from the three models. The missing values are shown in circles, while training points are shown with black dots. All models can accurately reproduce the overall trends, while npODE seems to fit slightly worse than the other methods. The PCA projection causes the seemingly perfect fit of the npODE prediction (at the top) to lead to slightly warped reconstructions (at the bottom). All methods mostly fit the missing parts as well. Table \ref{tab:t1} shows that on average the npODE and VGPLVM have approximately equal top performance on the imputing missing values task.


\section{Discussion}

We proposed the framework of nonparametric ODE model that can accurately learn arbitrary, nonlinear continuos-time dynamics from purely observational data without making assumptions of the underlying system dynamics. We demonstrated that the model excels at learning dynamics that can be forecasted into the future. We consider this work as the first in a line of studies of nonparametric ODE systems, and foresee several aspects as future work. Currently we do not handle non-stationary vector fields, that is time-dependent differentials $\f_t(\x)$. Furthermore, an interesting future avenue is the study of various vector field kernels, such as divergence-free, curl-free or spectral kernels \citep{remes2017}. Finally, including inputs or controls to the system would allow precise modelling in interactive settings, such as robotics.

The proposed nonparametric ODE model operates along a continuous-time trajectory, while dynamic models such as hidden Markov models or state-space models are restricted to discrete time steps. These models are unable to consider system state at arbitrary times, for instance, between two successive timepoints.

Conventional ODE models have also been considered from the stochastic perspective with stochastic differential equation (SDE) models that commonly model the system drift and diffusion processes separately leading to a distribution of trajectories $p(\x(t))$. As future work we will consider stochastic extensions of our nonparametric ODE model, as well as MCMC sampling of the inducing point posterior $p(U | Y)$, leading to trajectory distribution as well.

\paragraph{Acknowledgements.} 
The data used in this project was obtained from \url{mocap.cs.cmu.edu}. The database was created with funding from NSF EIA-0196217. This work has been supported by the Academy of Finland Center of Excellence in Systems Immunology and Physiology, the Academy of Finland grants no.~260403, 299915, 275537, 311584.


\bibliography{refs}
\bibliographystyle{plainnat}

\clearpage

\onecolumn

\section*{Appendix of `Learning unknown ODE models with Gaussian processes'}

\subsection*{Sensitivity Equations}

In the main text, the sensitivity equation is formulated using matrix notation
\begin{align}
    \dot{S}(t)  &= J(t)S(t) + R(t).
\end{align}
	Here, the time-dependent matrices are obtained by differentiating the vector valued functions with respect to vectors i.e.
	\begin{align}
		S(t) &= 
		\begin{bmatrix}
			\frac{dx_1(t,U)}{d u_1} & \frac{dx_1(t,U)}{d u_2}& \cdots & \frac{dx_1(t,U)}{d u_{MD}} \vspace*{0.2cm} \\
			\frac{dx_2(t,U)}{d u_1} & \frac{dx_2(t,U)}{d u_2}& \cdots & \frac{dx_2(t,U)}{d u_{MD}} \vspace*{0.2cm} \\
			\cdots & \cdots & \cdots & \cdots \\
			\cdots & \cdots & \cdots & \cdots \\
			\frac{dx_D(t,U)}{d u_1} & \frac{dx_D(t,U)}{d u_2}& \cdots & \frac{dx_D(t,U)}{d u_{MD}} \vspace*{0.2cm} 
		\end{bmatrix}_{D \times MD} \\ 
		J(t) &= 
		\begin{bmatrix}
			\frac{\partial f(\x(t),U)_1}{\partial x_1} & \frac{\partial f(\x(t),U)_1}{\partial x_2} & \cdots & \frac{\partial f(\x(t),U)_1}{\partial x_{D}} \vspace*{0.2cm} \\
			\frac{\partial f(\x(t),U)_2}{\partial x_1} & \frac{\partial f(\x(t),U)_2}{\partial x_2} & \cdots & \frac{\partial f(\x(t),U)_2}{\partial x_{D}} \vspace*{0.2cm} \\
			\cdots & \cdots & \cdots & \cdots \\
			\cdots & \cdots & \cdots & \cdots \\
			\frac{\partial f(\x(t),U)_D}{\partial x_1} & \frac{\partial f(\x(t),U)_D}{\partial x_2} & \cdots & \frac{\partial f(\x(t),U)_D}{\partial x_{D}}
		\end{bmatrix}_{D \times D} \\
		R(t) &= 
		\begin{bmatrix}
			\frac{\partial f(\x(t),U)_1}{\partial u_1} & \frac{\partial f(\x(t),U)_1}{\partial u_2} & \cdots & \frac{\partial f(\x(t),U)_1}{\partial u_{MD}} \vspace*{0.2cm} \\
			\frac{\partial f(\x(t),U)_2}{\partial u_1} & \frac{\partial f(\x(t),U)_2}{\partial u_2} & \cdots & \frac{\partial f(\x(t),U)_2}{\partial u_{MD}} \vspace*{0.2cm} \\
			\cdots & \cdots & \cdots & \cdots \\
			\cdots & \cdots & \cdots & \cdots \\
			\frac{\partial f(\x(t),U)_D}{\partial u_1} & \frac{\partial f(\x(t),U)_D}{\partial u_2} & \cdots & \frac{\partial f(\x(t),U)_D}{\partial u_{MD}}
		\end{bmatrix}_{D \times  MD}
	\end{align}

	\subsection*{Optimization}
	
	Below is the explicit form of the log posterior. Note that we introduce $\u = vec(U)$ and $\Omega = \diag( \omega_1^2, \ldots, \omega_D^2)$ for notational simplicity. 
	
	\begin{align}
		\log\L &= \log p(U|\bt) + \log p(Y | \x_0, U,\bo)\\
		&= \log\gN(\u|\mathbf{0},\K_\bt(Z,Z)) +  \sum_{i=1}^{N} \log \gN(\y_i|\x(t_i,U),\Omega) \\
		&= -\frac{1}{2} \u^T\K_\bt(Z,Z)^{-1}\u -\frac{1}{2}  \log |\K_\bt(Z,Z)| -\frac{1}{2} \sum_{i=1}^{N} \sum_{j=1}^{D} \frac{(y_{ij} - x_j(t_i,U,\x_0))^2}{\omega_j^2} -\sum_{i=1}^{N} \frac{1}{2}  \log |\Omega| \\
		&= -\frac{1}{2} \u^T\K_\bt(Z,Z)^{-1}\u -\frac{1}{2}  \log |\K_\bt(Z,Z)| -\frac{1}{2} \sum_{i=1}^{N} \sum_{j=1}^{D} \frac{(y_{i,j} -x_j(t_i,U,\x_0))^2}{\omega_j^2} -N\sum_{j=1}^{D} \log \omega_j
	\end{align}
	
	Our goal is to compute the gradients with respect to the initial state $\x_0$, latent vector field $\tU$, kernel parameters $\bt$ and noise variables $\bo$. As explained in the paper, we compute the gradient of the posterior with respect to inducing vectors $U$ and project them to the white domain thanks to noncentral parameterisation. The analytical forms of the partial derivatives are as follows:
	\begin{align}
		\frac{\partial \log\L}{\partial u_k} &= \sum_{i=1}^{N} \sum_{j=1}^{D} \frac{y_{i,j} -x_j(t_i,U,\x_0)}{\omega_j^2} \frac{\partial x_j(t_i,U,\x_0)}{\partial u_k} - \K_\bt(Z,Z)^{-1}\u \\
		\frac{\partial \log\L}{\partial (x_0)_d} &= \sum_{i=1}^{N} \sum_{j=1}^{D} \frac{y_{i,j} -x_j(t_i,U,\x_0)}{\omega_j^2} \frac{\partial x_j(t_i,U,\x_0)}{\partial (x_0)_d} \\
		\frac{\partial \log\L}{\partial \omega_j} &= \frac{1}{\omega_j^3} \sum_{i=1}^{N} (y_{i,j} -x_j(t_i,U,\x_0))^2 - \frac{N}{\omega_j}
	\end{align}
	
    Seemingly hard to compute terms, $\frac{\partial x_j(t_i,U,\x_0)}{\partial u_k}$ and $\frac{\partial x_j(t_i,U,\x_0)}{\partial (x_0)_d}$, are computed using sensitivities. The lengthscale parameter $\bl$ is considered as a model complexity parameter and is chosen from a grid using cross-validation. We furthermore need the gradient with respect to the other kernel variable, i.e., the signal variance $\sigma_f^2$. Because $\K_\bt(Z,Z)$ and $\x(t_i,U)$ are the functions of kernel, computing the gradients with respect to $\sigma_f^2$ is not trivial and we make use of finite differences:
	\begin{align}
	    \frac{\partial \log\L}{\partial \sigma_f} = \frac{\log\L(\sigma_f+\delta)-\log\L(\sigma_f)}{\delta}
	\end{align}
    We use $\delta = 10^{-4}$ to compute the finite differences.
 
	One problem of using gradient-based optimization techniques is that they do not ensure the positivity of the parameters being optimized. Therefore, we perform the optimization of the noise standard deviations ${\bo = (\omega_1, \ldots, \omega_D)}$ and signal variance $\sigma_f$ with respect to their logarithms:
	\begin{align}
	\frac{\partial \log\L}{\partial \log c} &= \frac{\partial \log\L}{\partial c} \frac{\partial c}{\partial \log c} = \frac{\partial \log\L}{\partial c} c
	\end{align}
	where $c \in (\sigma_f, \omega_1, \ldots, \omega_D)$.

	\subsection*{Implementation details}

    We initialise the inducing vectors $U = (\u_1, \ldots \u_M)$ by computing the empirical gradients $\dot{\y}_i = \y_i - \y_{i-1}$, and conditioning as
    \begin{align}
        U_0 = \K(Z,Y) \K(Y,Y)^{-1} c \dot{\y},
    \end{align}
    where we optimize the scale $c$ against the posterior. The whitened inducing vector is obtained as $\tU_0 = \bL_\bt^{-1} U_0$. This procedure produces initial vector fields that partially match the trajectory already. We then do 100 restarts of the optimization from random perturbations $\tU = \tU_0 + \varepsilon$.
    
    We use LBFGS gradient optimization routine in Matlab. We initialise the inducing vector locations $Z$ on a equidistant fixed grid on a box containing the observed points. We select the lengthscales $\ell_1, \ldots, \ell_D$ using cross-validation from values $\{0.5, 0.75, 1, 1.25, 1.5\}$. In general large lengthscales induce smoother models, while lower lengthscales cause overfitting.

\end{document}